# Diagnosis of Scalp Disorders using Machine Learning and Deep Learning Approach - A Review

Mr Hrishabh Tiwari
Computer Science
Dr Vishwanath Karad, MIT World Peace Universiy
Pune, India
1032201436@mitwpu.edu.in

Mr Jatin Moolchandani
Computer Science
Dr Vishwanath Karad, MIT World Peace Universiy
Pune, India
1032201427@mitwpu.edu.in

Dr Shamla Mantri
Computer Science
Dr Vishwanath Karad, MIT World Peace Universiy
Pune, India
shamla.mantri@mitwpu.edu.in

*Abstract*— The morbidity of scalp diseases is minuscule compared to other diseases, but the impact on the patient's life is enormous. It is common for people to experience scalp problems that include Dandruff, Psoriasis, Tinea-Capitis, Alopecia and Atopic-Dermatitis. In accordance with WHO research, approximately 70% of adults have problems with their scalp. It has been demonstrated in descriptive research that hair quality is impaired by impaired scalp, but these impacts are reversible with early diagnosis and treatment. Deep Learning advances have demonstrated the effectiveness of CNN paired with FCN in diagnosing scalp and skin disorders. In one proposed Deep-Learning-based scalp inspection and diagnosis system, an imaging microscope and a trained model are combined with an app that classifies scalp disorders accurately with an average precision of 97.41%-99.09%. Another research dealt with classifying the Psoriasis using the CNN with an accuracy of 82.9%. As part of another study, an ML based algorithm was also employed. It accurately classified the healthy scalp and alopecia areata with 91.4% and 88.9% accuracy with SVM and KNN algorithms. Using deep learning models to diagnose scalp related diseases has improved due to advancements in computation capabilities and computer vision, but there remains a wide horizon for further improvements.

## I. Introduction

Hair is a skin appendage that shares a common pathway of development with other ectodermal tissue [17]. An individual's hair plays an instrumental role in defining their identity, social status, and level of confidence [9]. Considering the current environmental scenario, hair nourishment and optimal development is a very difficult task. Stress and dietary changes cause a significant increase in hair-loss, both in men and women [14]. People often overlook the early signs of scalp diseases that excavate hair loss [8]. Endocrine, genetic, chronic disease, and other internal factors are some reasons for scalp and hair problems [1,4,13]. Seborrheic dermatitis, folliculitis, psoriasis, tinea capitis, alopecia, and atopic dermatitis are some of the frequently observed scalp diseases [10,22].

Chronic and recurrent, seborrheic dermatitis affects skin causing dandruff and scaling in areas rich in sebaceous glands [18]. There are two such stages in the human life cycle, the first occurring during adolescence and the second during puberty and adulthood [18]. Research has shown that excessive growth of Malassezia (type of Fungus), androgens, and skin lipids on the scalp are the primary causes of the disorder [6,18]. Additionally, seborrheic dermatitis can be caused by a variety of risk factors, such as age, male sex, immunodeficiency (like HIV-AIDS, renal transplantation, lymphoma), neurological and psychiatric disorders (such as Parkinson's disease, Alzheimer's dementia, stroke, major depression, and autonomic dysfunction), and drug treatment involving dopamine antagonists, immunosuppressants, lithium and psoralen [18].

There are bald patches on the forehead associated with Alopecia Areata, an autoimmune hair disorder [5,6,19]. This autoimmune disease is caused by clusters of T-lymphocytes coordinated at the hair follicle to cause this condition [3,5,23]. Reports have observed that people suffering from severe Alopecia areata have family members suffering from Asthma, Down Syndrome, Pernicious Anemia and Thyroid Disease etc. Researchers have surveyed that there are more than 1 million cases observed in India annually and are showing phenomenal growth [2]. However, with limited medical professionals and unreliable treatments available in the market, presents tremendous challenges to control the outbreak [5,23].

Psoriasis is a prevalent scalp disease-causing, red-skinned plaques on the scalp [5,16]. WHO reports more than 125 million cases annually, making it a global problem [5]. The reason behind the occurrence of Psoriasis is unknown, however, researchers suggest that excessive smoking, drinking, and mental anxiety [16,22]. This triggers uncontrollable growth of new skin cells which are replaced in 3 to 4 days rather than the normal replacement period of 10-15 days. This creates silver color patches on scalps and certain appendages of the skin. Psoriasis condition has been differentiated into 5 different sets which include Guttate psoriasis, Plaque psoriasis, Erythrodermic psoriasis, Pustular psoriasis, and Inverse psoriasis [16]. In the following study we will be extensively analyzing the classification and diagnosis of scalp Psoriasis.

Traditionally, scalp conditions such as Seborrheic Dermatitis, Psoriasis, Tinea Capitis, Folliculitis, Alopecia Areata, Tinea

Ringworm, Eczema, and Telogen Effluvium are diagnosed with biopsy procedures [10,20]. Biopsies involve the removal of small tissue cells from the scalp for laboratory analysis. Besides being complicated, this procedure is excruciating for patients as well. Therefore, the biopsy procedure is dependent on the physiotherapist's level of experience as to how they interpret scalp hair microscopy images [1].

To overcome the above-mentioned problems, medical sciences have opted for Deep Learning technology. Deep Learning is a branch of machine learning that functions upon multiple cascade layer algorithmic architecture for analyzing data. Data is filtered through multiple cascade layers, with each successive layer using the output from the previous one to form its results. The following text will explore a variety of innovations in the field of scalp disorders and classification of scalp diseases using different aspects of machine learning and deep learning.

## II. Literature Review

Hair Loss is the topic of enormous public interest and understanding the hair anatomy to comprehend the classification of hair disorder will bring a significant impact on humankind [25]. Our community suffers from a variety of hair disorders such as Seborrheic Dermatitis, Alopecia Areata, Eczema, Androgenetic Alopecia, Psoriasis, and other genetic disorders. Trichologists and medical researchers have been rigorously working upon distinctive hair disorders to find a common solution for diagnosis. Subsequently, it has been proven that deep learning has the potential to generalize these disorders for further analysis.

The system proposed by Wan-Jung Chang et al. is called ScalpEye, which was based on RCNN (Region-Based Convolutional Neural Network) using the ResNet_v2_Atrous model [1,33]. An imaging microscope on a portable device, a mobile application, a cloud-based artificial intelligence training server, and a management platform were integrated into the approach. The system was able to evaluate and diagnose four common types of disorders which are folliculitis, oily hair, dandruff, and hair loss. The Average Precision ranges from 97.41% to 99.09%, making it one of the most accurate systematic approaches for hair disorder classification. Choudhary Sobhan Shakeel et al. [2] on the other had taken a distinctive course of action, where the author focused on distinguishing between healthy hairs and scalps suffering from alopecia areata. Their Framework used a k-nearest neighbor (KNN) and support vector machine (SVM) which encompasses the color, textures, and shape of the subject's hair for classification. However, the precision for the following algorithms were 91.4% and 88.9% respectively. Hence, proving to be a successful framework for practical application.

As part of a study by Gregor Urban et al., OCT imaging has been combined with Deep Learning to calculate the number of hair follicles and hair density of the scalp [9]. Optical Coherence Tomography is a non-invasive process that uses infrared light on skin tissue benefiting from its interference properties. It can evaluate and analyze the morphology of the skin to induce hair growth. The author also asserted that the precision/accuracy of the framework is within the limits of the human discrepancy range. Skin in the Human body acts as a shield against various fungal, and bacterial infections and controls the temperature of the body [16].

There are several skin diseases, some are very common while some are rare. According to statistics, around 125 million people from all over the world suffer from different types of skin infections [16]. Rosniza Roslan et al. [16] have also introduced a non-invasive method for detecting psoriasis disease. Using the same CNN technique, the author was able to classify a variety of psoriasis, for further diagnosis [7]. Using CNN to diagnose skin diseases accurately has shown promising results [21,42]. Precision and accuracy were asserted to be exceptional, making it an ideal psoriasis detection technique. A more innovative approach to dermoscopic image analysis was proposed by Meng Gao et al. in which multiple ordinal logistic regression was used to predict basic and specific classification [12].

Masum Shah Junayed et al. [14] has proposed a deep CNN-based Eczema classification system, known as EczemaNet. In this framework, the author classifies 5 types of Eczema using deep-CNN and data augmentation to transform images for better results. With the help of regularization techniques (including batch normalization and dropout), the approach was able to achieve an accuracy of over 96.2%. A comparative study between the MobileNetV1 pre-trained models and EczemaNet model with InceptionV3 was thoroughly analyzed with custom datasets of 2000 images of 5 different types of Eczema classes. Which concluded with EczemaNet surpassing the other two pre-trained models. For future development, the author suggested implementing Segmentation and accurate detection techniques, also by increasing the amount of data per class to achieve better accuracy.

The work of Anastasia Butskova et al. introduces several aspects of Deep Learning employing multiple algorithms, including mix-up data augmentation, self-supervised learning, and two stages EfficientNet with image context disordering [24]. According to the experiment's conclusion, mix-ups and focal losses that have a positive gamma hyperparameter improve the average recall, regardless of the class of mix-up.

## III. Methodology

Table 1: Methods, Datasets, and the performance of different studies in diagnosis of scalp disorders. [1,2,15,16]

| Sr No. | Research Paper Findings | Method | Dataset | mAP |
|---|---|---|---|---|
| 1. | Scalp hair inspection and diagnosis based on deep learning [1] | DL approach that classifies 4 types of scalp conditions using CNN | 2,000 scalp hair microscopic images collected from Taiwan Hair Dressing Industry that included four different sets of scalp symptoms. | 97.41% - 99.09% |
| 2. | Eczema Classification [15] | A DL approach based on CNN and computer vision to classify multiple types of skin eczema. | 100 images were collected from hospitals, 400 images from open-source websites. | 96.2% |
| 3. | Psoriasis Classification [16] | Diagnosis of Psoriasis using Convolutional Neural Network. | A total of 187 images consisting of 105 images for Guttate Psoriasis and 82 images for Plaque Psoriasis | 72.4% - 82.9% |
| 4. | Alopecia Areata Classification [2] | An ML based approach that using KNN, SVM and Decision Tree models to classify healthy scalp and Alopecia Areata condition by analyzing hair. | 68 hair images with alopecia areata symptoms from the Dermnet dataset and 200 images of healthy hairs from the Figaro 1k dataset | 88.9% - 91.4% |

### A. Research Method

The objective of the proposed system by Wan-Jung Chang et al. [1] was to diagnose and classify different types of scalp conditions such as Folliculitis, Dandruff, Hair Loss, and Oily Hair [1]. Existing research focused on the diagnosis of a single class of scalp disorder. In this particular study, the authors added to the previous works, and thus developed a system capable of diagnosing and classifying different types of scalp disorders. This research work is very stately since the Deep Learning Model built with the help of computer vision and ConvNet is portable, efficient and has provided high accuracy.

Alopecia Areata is one of the common types of scalp disorders that leads to patchy hair loss, mostly from the crown and the temporal regions of the scalp [11,26]. Machine Learning has demonstrated promising potential for the diagnosis of various scalp related disorders including alopecia areata [2]. In the study to diagnose alopecia areata, authors Choudhary et al. [2] primarily implemented two machine learning algorithms known as k-nearest neighbor (KNN) and support vector machine (SVM) for the identification and diagnosis of the auto-immune disease. Former researchers have emphasized dermoscopic and scalp images using only the SVM algorithm. The study, however, did not just include shape, color, and texture as features, but also SVM and KNN as algorithms for classification.

Eczema is the most common skin disease and traditionally the diagnosis of eczema is done by dermatologists or doctors [15,30]. To help the patients diagnose the skin disorder themselves, Masum Shah et al. have proposed a deep learning approach. The author's study was designed to classify five different types of Eczema. The study did not limit itself to just the scalp but provided a method to classify eczema from all parts of the body [15]. Authors Masum Shah et al. proposed a deep CNN to extract the feature map from the image. With an adapted Fully Connected Neural Network, the five types of Eczema viz. Subacute Eczema, Hand Eczema, Nummular Eczema, Chronic Eczema, and Asteatotic Eczema were identified [15].

It has been estimated that in a sample size of 296 Psoriasis patients, around 59% are from Malaysia and 28% are from the Indian subcontinent. From the numbers, both these nations have a population prone to psoriasis. The high amount of population suffering from challenges remains the correct diagnosis of the type of skin disease for medications. To overcome this challenge, Rosniza Roslan et al. [16], in their study have proposed a Deep Neural Networks Architecture to assort different types of Psoriasis disorders. Using Convolutional Neural Network for extracting key features from the image, the proposed model could identify two types of Psoriasis condition i.e., Plaque and Guttate with an accuracy of 82.9% and 72.4%.

### B. Datasets

The lack of available datasets makes medical diagnosis very challenging. To build a Deep Learning model with high accuracy and high precision, a good amount of data is required to be processed [28]. In the scalp diagnosis model of ScalpEye, almost 2000 hair scalp images from Taiwan Hairdressing companies were obtained for classification [1]. Since the dataset was obtained from multiple sources, it needed a lot of preprocessing before any kind of Neural Architecture could be trained.

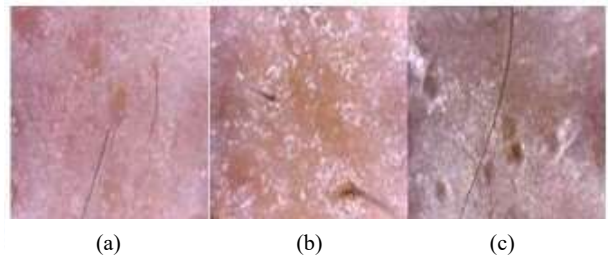

(a)      (b)      (c)

*Fig. 1 Dandruff Symptoms: a) Normal b) Middle c) High [1]*

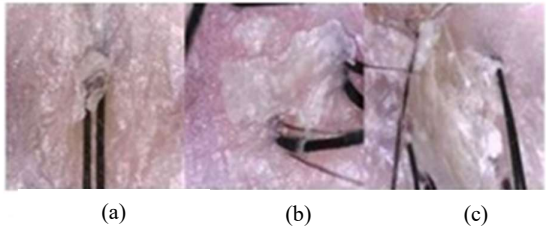

*Fig.2 Hair loss Symptoms: a) Normal b) Middle c) High [1]*

Figure 1 and 2 shows some of the sample images that were used by Wang-Jung et al. for developing the ScalpEye Pipeline [1]. To annotate the dataset images with bounding boxes for model training, extensive help was taken from medical experts and physiotherapists.

Medical diagnosis using KNN and SVM requires a considerable amount of dataset for better accuracy and precision of the model. In the following research, the authors Choudhary et al. [2] have used more than 68 images of scalps from the Dermnet Dataset. 200 images from the Figaro1k Dataset were also used by the authors Choudhary et al. [2] aimed at classifying Alopecia Areata based on the Trichoscopic features. The Figaro1k dataset is an open-source dataset that contains different classes of hair such as curly, wavy, straight etc. [32]. The 268 images are very little data for training a Neural Network hence Choudhary et al. [2] have taken the Machine Learning Algorithmic approach to classify Alopecia Areata. Researchers typically use scalp images to analyze skin features associated with alopecia areata [2]. Using encapsulated techniques, such as eigenvalues, and grid line selection, a Trichoscopic method was proposed by Choudhary et al. that extracted hair loss features from scalp images. Figure 3 shows some of the images from the dataset used for classification of Alopecia Areata by Choudhary et al [2].

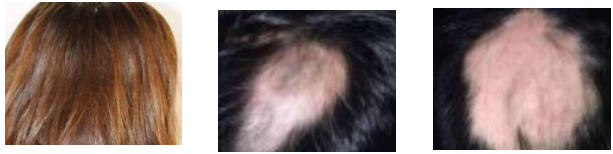

*Fig.3 Sample Images from the dataset used by Choudhary et al. [2]*

For the eczema classification, 500 images of different type of eczemas were used, 400 of them were obtained from open sources while 100 were collected from different Hospitals. As the number of images were very low to train a deep neural network and obtain good performance, a lot of hand engineering was involved. The dataset was augmented with different kinds of augmentation such as flipping, rotation, shearing, zoom, translation etc. The augmentation not only results in creation of more data from existing data but also results in training the Deep Neural Networks model with a real life-like dataset which would be encountered by the model at the time of inference.

*C. Models*

To determine the correct Deep learning model for implementing a medical diagnosis is somewhat a challenging task [1]. Wang-Jung et al [1] used the following 3 models during the research:

a.) FASTER R-CNN INCEPTION_V2
b.) SSD INCEPTION_V2
c.) FASTER R-CNN INCEPTION_RESNET_V2_ATROUS

The following models were chosen after carefully examining the mAP (Mean Average Precision) and the speed of the deep learning models on the COCO Dataset [34]. The COCO Dataset contains 328k images of 91 different classes and has over 2.5 million labeled instances. The dataset is open source and a benchmark for object detection and segmentation models [34].

Faster R-CNN Inception_V2 model used by Wang-Jung et al. [1] is based on the Faster-RCNN technique for object detection. RPN illustrated in the figure 4 is used as a region proposal algorithm, and Fast R-CNN is used as a detector network in the Faster R-CNN architecture shown in the figure 5 [31]. The CNN trains independently from the given dataset and can identify specific objects in an image using machine learning algorithms [37,40]. In Faster R-CNN, regions are proposed for extracting features from an image. Region Proposal Networks (RPNs) take images (of any size) as input and produce rectangular object proposals, each with its score of objectness [31]. Using the backbone layer's convolutional feature map as input, this region proposal network generates anchors by applying sliding window convolution to the input map. To extract the feature map, Wang-Jung et al. used the Inception_V2 architecture as the backbone [1].

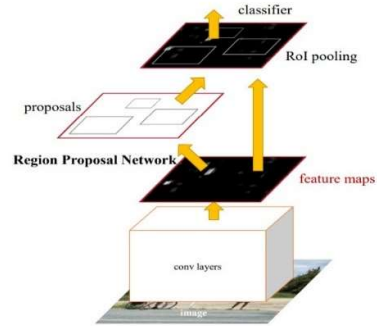

*Fig.4 Region Proposal Network Module [31]*

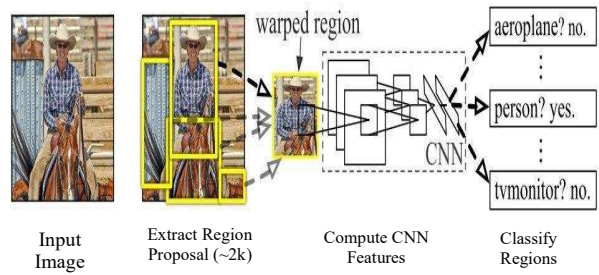

*Fig.5 A high-level view of Faster R-CNN Architecture [31]*

**SSD Inception_V2:** SSD stands for Single Shot Detector, this model paired with Inception V2 provided many folds speed gains over the Faster RCNN model. The Faster R-CNN model first proposes some 'Regions of Interests' and then classifies those regions based on some convolutional and fully connected layers whereas the Single Shot Detector does both the things in one go. The SSD works in a fashion where the Input image passed through Convolutional layers and it extracts feature maps of different sizes (10X10, 6X6, 3X3 etc.) and using these feature maps it evaluates a small set of pre-defined bounding boxes [41]. For each box, it predicts a class probability, and the box coordinates like the Faster R-CNN. One of the least complicated versions of an inception module works by combining three different filter sizes into one input with different kernel sizes i.e., 1x1, 3x3, 5x5.

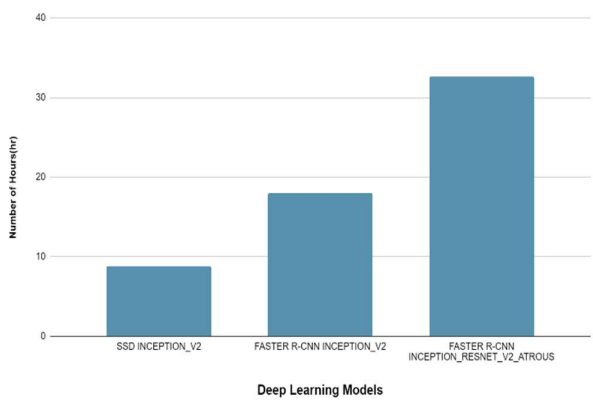

*Fig.6 Deep Learning Models used by Wang-Jung et al. for diagnosis of various scalp diseases and their respective training time in hours [1]*

To accurately detect and diagnose the scalp disorder, image preprocessing is required for improving the pixel luminance, contrast, and brightness value using Histogram equalization [2]. Consequently, the images are also refined using image segmentation and edge detection. Several preprocessing steps precede the classification process, which involves the extraction of three features: the hair color, the texture, and the shape. The Histogram Equalization method has been used for evenly distributing the high intensity over a wide region in the image [29]. When a narrow range of intensity values represents an image, the Histogram Equalization method typically increases its global contrast. The intensity range can be more evenly distributed on the histogram using this adjustment. High contrast can be achieved in areas with a low local contrast because of this.

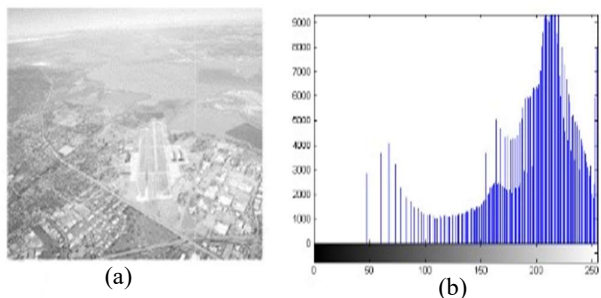

(a)　　　　　　　　　(b)

*Fig.7 (a) Low contrast in Image (b) Pixel intensity Histogram in image [29]*

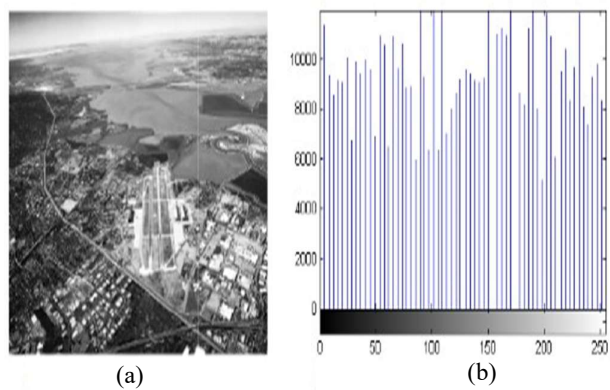

(a)　　　　　　　　　(b)

*Fig.8 Even contrast in Image (a) and its Pixel intensity Histogram in image (b) after HE. [29]*

Figures 7 and 8 illustrate the use of Histogram Equalization to evenly distribute the contrast and brightness values. With the help of CV2 and Skimage python libraries, segmented images are then processed for extracting various features such as color texture and shape.

**1. Color feature extraction**: The images are converted to array using color pixel values of red, green, and blue respectively, which are then averaged and packed using CV2 library.

**2. Texture Feature:** The extraction of these features is implemented using skimage and the cv2 library (imported to Scikit Learn using python). Local binary pattern is used as an algorithm to compute the descriptors of the texture.

**3. Shape Feature**: The extraction of this feature is done using OpenCV python library

**Support-vector Machine (SVM):** The objective of this Machine learning algorithm is to find a hyperplane that distinctively classifies the given data points [2]. When we consider data points as a p-dimensional vector (a list of p numbers), we can separate them with a (p-1) dimensional hyperplane using support vector machine algorithm. It is possible to classify the data using many hyperplanes. An appropriate method of selecting a hyperplane is to choose the one representing the greatest separation between the two classes, or margin. As a result, the hyperplane should be chosen so that it is the closest on either side to each data point. As long as such hyperplanes exist, they are called maximum-margin hyperplanes and their linear classifiers are called maximum-margin classifiers [27,35].

The results of the Support Vector Machine algorithm can be boosted with the use of different Kernel Tricks. The Kernel Tricks is a method wherein linear data is projected onto higher dimensions making it easy for the separation of data points using hyperplanes. Various Kernel Functions such as Gaussian RBF, Polynomial, Sigmoid, Linear, Hyperbolic kernel functions. In the Alopecia Areata classification, the radial basis function (RBF) has been used by Choudhary et al. [2].

$$\max_{\alpha} \sum_{i=1}^{m} \alpha_j - \frac{1}{2} \sum_{i=1}^{m} \sum_{j=1}^{m} \alpha_i \alpha_j y_i y_j (x_i x_j) \quad (1)$$

Eq. 1 The Kernel Trick as used by Choudhary et al. [2]

$$k(x,y) = e^{-\gamma \|x-y\|^2}, \gamma > 0 \quad (2)$$

Eq. 2 Gaussian RBF equation as used by Choudhary et al. [2]

**KNN (k-Nearest Neighbors):** K-nearest neighbor classification, a conventional non-parametric classifier, is used to solve a wide variety of pattern classification problems. A distance is measured between the test data and each training data to determine the final classification result. In the case of continuous data, Euclidean distance is used to calculate its nearest neighbors. Based on the calculated value of K, nearest neighbors determine the classification for new inputs. In spite of the simplicity of this classifier, K has a significant impact on the classification of unlabeled data. Calculations are made when classifying the training data, not when it appears in the dataset, so there is a slight computation cost involved. Since the training data is only stored and memorized when the dataset is being trained, the algorithm is one of the easiest and most powerful [28].

$$dist\big((x,a)(y,b)\big) = \sqrt{(x-a)^2 + (y-b)^2} \quad (3)$$

Eq. 3 Euclidean Distance

The authors Mausam Shah et al. [2] proposed an architecture having 3 Convolutional layers with the kernel size of 3X3. Three Max Pooling Layers were incorporated after each Convolutional Layer to decrease the dimensions of the image, one flatten layer was added after the last max pooling layer, the flatten layer was then followed by three Fully Connected or Dense Layers. Rectified Linear Units also known as ReLU has been used as an activation function with the Convolutional Layers.

ReLU are one of the widely used activation functions which simply clamp the negative values to zero and keep the positive values as they are [36]. This activation function is helpful when predicting the bounding boxes for object localization and detection tasks. Batch Normalization technique has also been used after the last two Max Pooling layers and after each fully connected layer. By converting the joint feature representation into a pooling layer, CNNs can keep the relevant information while removing unnecessary details [39]. The Batch Normalization helps in smoother convergence and does help in curving down the gradient explosions [38]. Apart from this, regularization has also been used in the architecture of the model to reduce the risk of overfitting, the regularization functions such as L1 or L2 also known as Lasso and Ridge regularizers (equation 4, 5) have proven to increase the overall performance of the model at the time of inference on data that have never been seen before by the model. Furthermore, the authors Mausam Shah et al.

have used the SoftMax activation in the output of the architecture. The SoftMax function (equation 6) gives probability of each class in the given input.

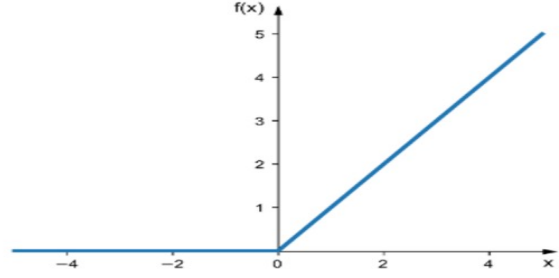

Fig. 9 Graph of the ReLU function [36]

$$m \sum_{j=1}^{m} |\hat{\beta}_j| \quad (4)$$

Eq. 4 Lasso or L1 Regularization Term

$$m \sum_{j=1}^{m} |\hat{\beta}_j|^2 \quad (5)$$

Eq. 5 Ridge or L2 Regularization Term

$$\frac{exp(x_i)}{\sum_j exp(x_j)} \quad (6)$$

Eq. 6 SoftMax Function

The Neural Architecture utilized by Mausam Shah et al. [2] is quite a standard one in practice. The model architecture is neither too deep nor too shallow but considerable with not too much complexity involved. The deep learning model was trained for 60 epochs/iterations with a batch size of 32 that yielded an accuracy of 96.2%. Two more models InceptionV3 and MobileNetV1 were trained using the technique of transfer learning. In Transfer Learning, pretrained models are used and fine-tuned for custom needs, the transfer learning method is widely used since it provides a head-start for the training, it is also useful when there is a limited amount of data available for training a complex model.

The dataset for Psoriasis disease classification has been obtained from International Psoriasis Council, Psoriasis Image Library, and the DemaNet NZ website. For the classification of the Psoriasis into two categories i.e., Plaque and Guetta, a total of 187 images have been used. Out of these 187 images, 82 images were of Plaque Psoriasis and 105 images were of Guttate Psoriasis. The research for the classification of Psoriasis is not limited to the scalp but the authors Rosniza et al. [16] have proposed a framework to diagnose it from all over the body. Since the number of images were fairly low and the dataset was obtained from various source, a lot of preprocessing was required to get the best out of the Convolutional Neural Networks. CNNs form the core when it comes to computer vision, they almost every time outperform Fully Connected Network. Rosniza et al. have used a network similar to AlexNet architecture [16]. AlexNet had eight layers; among the eight layers, five were convolutional layers, some of these Conv Layers layers were followed by max-pooling layers. The AlexNet also incorporates three were fully connected layers also known as the Dense layers where each neuron is connected to the other. In comparison with tanh and sigmoid, the non-saturating ReLU activation function used as the non-linear activation function showed better training performance.

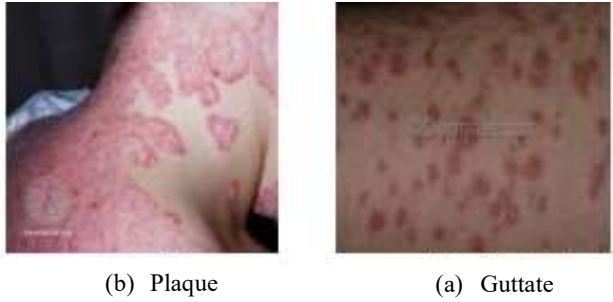

(b) Plaque      (a) Guttate

*Fig. 10 Sample images from the datasets used by Rosniza et al. [16]*

### D. Results

Table 2: Deep Learning Models used for the diagnosis of Eczema with their performance metrics [15]

| Sr No. | Model | Accuracy | Precision |
|---|---|---|---|
| 1 | InceptionV3 | 91% | 87% |
| 2 | MobileNetV1 | 92% | 88% |
| 3 | EczemaNet | 96.2% | 90% |

The custom-made Architecture EczemaNet yielded the best results for the classification of five different types of Eczema [15]. The accuracy of the model was found out to be 96.2% with a precision of 90%. The InceptionV3 pretrained model's accuracy was observed to be 91% with a precision of 87% while the MobileNetV1 yielded an accuracy of 92%. It is clear from the performance evaluation that the less complex model with few ConvNet outperformed other models.

Table 3: ScalpEye Deep Learning-Based Scalp Hair Inspection and Diagnosis System for Scalp Health [1]

| Disorder→<br>Model ↓ | Dandruff | Folliculitis | Hair Loss | Oily Hair |
|---|---|---|---|---|
| Faster R-CNN Inception_V2 | 96.44% | 92.62% | 99.95% | 92.60% |
| SSD Inception_V2 | 91.34% | 88.24% | 87.86% | 86.44% |
| Faster R-CNN INCEPTION_RESNET_V2_Atrous | 97.41% | 97.61% | 99.09% | 97.69% |

The following models were successfully implemented with Mean Average precision for diagnosing Dandruff as 96.44%, 91.34%, and 97.41% respectively. For Folliculitis, they were observed to be 92.65%, 88.24%, and 97.61% respectively. For Hair-Loss they were 99.95%, 87.86% and 99.09% respectively and for analyzing Oily hairs they were observed to be 92.60%, 86.44%, and 97.69%. The results clearly indicated that the Faster R-CNN Inception-ResNet v2-Atrous model yields better performance with a reasonable computation and processing time.

The Support Vector Machine and the k-Nearest Neighbors were trained on a set of 187 images out of 268 images present and the trained algorithms were evaluated by using 81 test images. The Support Vector Machine outperformed the k-Nearest Neighbors algorithm with an accuracy of 91.4% while KNN had an accuracy of 88.9%. The algorithms were evaluated with the technique of k-Fold validations. The SVM had a precision of 96.3% while the KNN had a precision of 100% for positive classes i.e., images with Alopecia Areata. The F1 score for the Alopecia Areata class was observed to be 93.7% and 91.4% with SVM and KNN Algorithms respectively. It is true that the accuracy and F1 score that provide a good metric for evaluating the algorithms could be improved, but given the limited data used to train the framework, the algorithms have excelled. There is scope for major improvements, increasing the dataset for training is very likely to boost the performance of the algorithms. Nevertheless, an attempt to classify Alopecia Areata was done using hair features rather than scalp which produced considerable results.

Table 4: Performance of the ML Algorithm Framework used by Choudhary et al. for the classification of Alopecia Areata [2]

| Sr No. | ML Algorithm | Prediction Class | Accuracy | Precision | Recall | F1 Score |
|---|---|---|---|---|---|---|
| 1 | SVM | Non-Alopecia Areata | 91.4% | 81.5% | 91.7% | 86.3% |
| 2 | SVM | Alopecia Areata | 91.4% | 96.3% | 91.2% | 93.7% |
| 3 | KNN | Non-Alopecia Areata | 88.9% | 72.7% | 100% | 84.2% |
| 4 | KNN | Alopecia Areata | 88.9% | 100% | 84.2% | 91.4% |

Since the amount of data used to train the architecture was very low, it was quite hard to extract a good accuracy and it is apparent from the performance of the model. The framework had an accuracy of 82.9% for diagnosing the Psoriasis in the Plaque category whereas an accuracy of only 72.4% was observed for the classification of the Guttate Psoriasis. The model yielded precision of 82.9% and F1 Score of 75.98%.

### E. Conclusion and Future Work

Two Machine Learning Algorithms and several Deep Learning Models like AlexNet, Faster R-CNN, SSD MobileNetV1 etc. have been used for the diagnosis of several skin and scalp diseases. Dataset has been obtained from various sources, hence, to achieve a good accuracy, a lot of hand engineering and techniques of preprocessing and data augmentation has been used.

Amongst the DL frameworks, Convolutional Layers have been a common facet which has been utilized to extract key feature maps from the images. While the Machine Learning Algorithms like Support Vector Machine and k-Nearest Neighbor have obtained a good accuracy, they could have achieved even better results if a Convolutional Neural Network was utilized since the ConvNet are known to work better with the images. Faster R-CNN and SSD MobileNetV1 used in the ScalpEye [1] pipeline remain the standout performers among all the architectures diagnosing several types of scalp diseases.

Most of the models that yielded a better performance were trained on a dataset having at least 500 images for more than 8 hours. The future research works could take inspiration from the architectures like ScalpEye [1] and use a larger dataset to achieve even better accuracy. Research work in the domain of skin-diseased image enhancement and detailing could also be done to make the convolutional neural networks to extract even detailed feature maps. The Trichoscopic data such as hair density, thickness, color could be used to find the relevance and impact of various scalp diseases with the hair features.